\documentclass[letterpaper, 10 pt, conference]{ieeeconf}  

\IEEEoverridecommandlockouts                              

\usepackage{graphics} 
\usepackage{epsfig} 
\usepackage{amsmath} 
\usepackage{amssymb}  

\title{\LARGE \bf
Efficient Object Detection for High Resolution Images
}

\author{Yongxi Lu$^{1}$ and Tara Javidi$^{1}$
\thanks{$^{1}$Yongxi Lu and Tara Javidi are with the Department of Electrical and Computer Engineering, University of California, San Diego, La Jolla, CA 92093, USA.
        {\tt\small \{yol070, tjavidi\}@ucsd.edu}}%
}

\begin{document}

\maketitle
\thispagestyle{plain}
\pagestyle{plain}

\begin{abstract}

Efficient generation of high-quality object proposals is an essential step in state-of-the-art object detection systems based on deep convolutional neural networks (DCNN) features. Current object proposal algorithms are computationally inefficient in processing high resolution images containing small objects, which makes them the bottleneck in object detection systems. In this paper we present effective methods to detect objects for high resolution images. We combine two complementary strategies. The first approach is to predict bounding boxes based on adjacent visual features. The second approach uses high level image features to guide a two-step search process that adaptively focuses on regions that are likely to contain small objects. We extract features required for the two strategies by utilizing a pre-trained DCNN model known as AlexNet. We demonstrate the effectiveness of our algorithm by showing its performance on a high-resolution image subset of the SUN 2012 object detection dataset. 

\end{abstract}

\section{INTRODUCTION}

The recent rapid developments of visual recognition systems are driven by three factors: the adoption of architectures based on deep convolutional neural networks (DCNN), the availability of large datasets such as SUN \cite{xiao2010sun}, Pascal VOC \cite{everingham2010pascal} and ImageNet \cite{ILSVRC15}, and the developments of high performance parallel computations. A landmark in this wave of increased recognition accuracy is the AlexNet \cite{krizhevsky2012imagenet} model, which showed far superior performance in the challenging 1000-class ImageNet object classification task to previous approaches and consequently brought about a paradigm shift in the domain of object recognition.

This breakthrough in object classification has since inspired researchers to tackle the more challenging task of object detection \cite{szegedy2013deep}\cite{sermanet2013overfeat}\cite{girshick2014rich}. R-CNN algorithm \cite{girshick2014rich}, VGG \cite{simonyan2014very} and GoogLeNet \cite{szegedy2014going}\cite{szegedy2014scalable} introduced a clear framework that connects the task of object classification to that of object detection. Note that the improved performance of VGG and GoogLeNet are primarily due to an adoption of deeper DCNN and larger datasets.

As shown in \cite{girshick2014rich}\cite{simonyan2014very}\cite{szegedy2014going}\cite{szegedy2014scalable}, the task of object detection intrinsically benefits from developments of accurate object classification. In the most elementary form, an object detection algorithm can: 1) Produce bounding boxes in an image as proposals for the object classification. 2) Each bounding box is then classified accurately via a DCNN. In other words, the parallel application of an accurate classifier to a set of bounding boxes of different sizes, locations and aspect ratios can be viewed as a basic object detection algorithm whose accuracy and performance significantly benefit from that of the object classifier used.

As a first step, the sliding window search scheme \cite{dollar2012pedestrian}\cite{felzenszwalb2010object} can be combined with a DCNN classifier to arrive at a set of bounding boxes of interest for an object detection task. Sliding window search, however, produces an excessive number of windows to be classified. Although DCNN models benefit from GPU acceleration, this simple approach, based on classifying tens of thousands of windows, fails to scale even for small to moderate size images. Authors in \cite{girshick2014rich}, instead, propose R-CNN in which the sliding window search is replaced by a fast pruning step known as object proposal generation using the selective search \cite{uijlings2013selective} algorithm. This step, in effect, restricts the extraction of deep features to around 2000 boxes per image. This algorithm can process a modest size image (a.k.a around $500 \times 500$) in around 2s. The computational bottleneck of R-CNN lies in the extraction of deep features of every of these roughly 2000 object proposals, which takes around 10-20s per image. 

\begin{figure*}[thpb]
      \centering
      \parbox{7in}{\includegraphics[height=3in, width=7in]{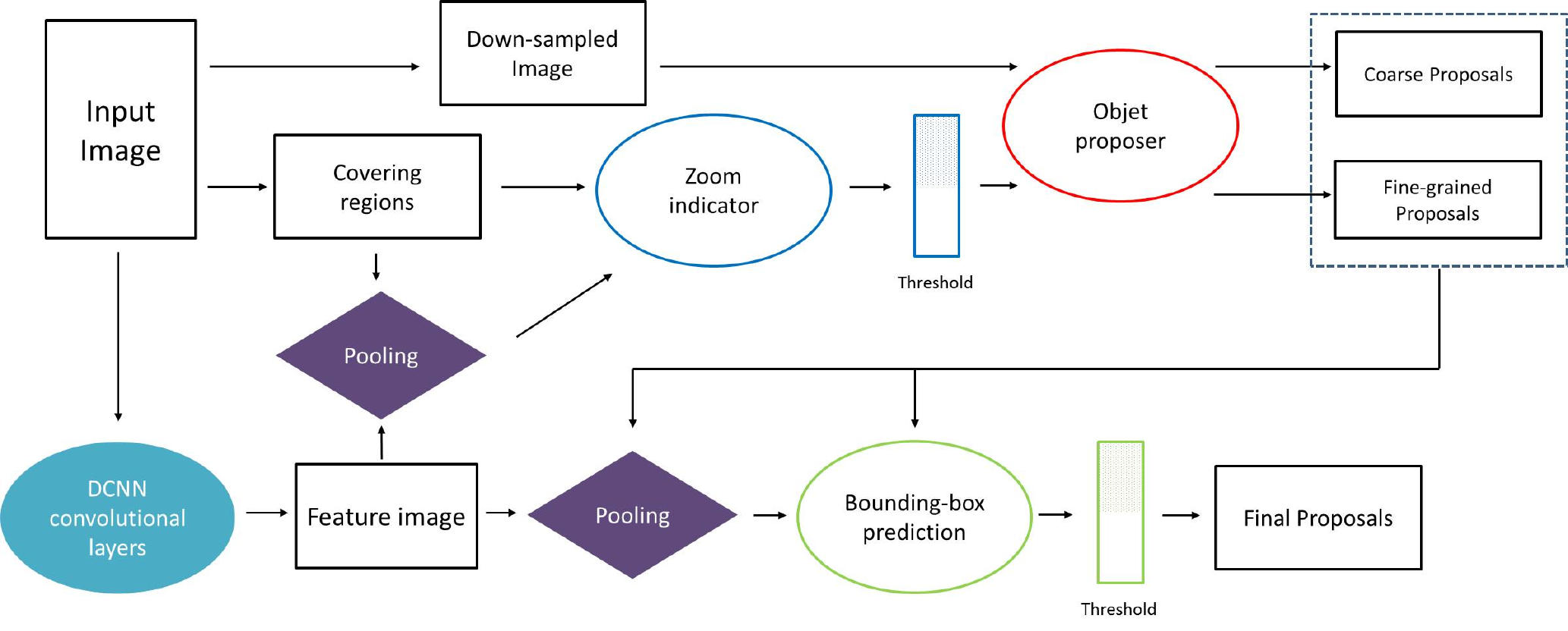}
      }
      \caption{The block diagram of our algorithm. At its input is an input image, and at its output is a set of bounding boxes called final proposals. Note that for a complete object detection pipeline a per-box classifier is applied at each final proposal, and a post-processing step follows. Our main focus is the steps leading to the final proposals.}
      \label{fig:pipeline_rough}
   \end{figure*}

Recent introduction of fast R-CNN \cite{girshick2015fast} have significantly improved on the run time and accuracy of the DCNN-based object detection algorithms by going beyond this two step approach. Instead of feature extraction and classification of each object proposal, in fast R-CNN \cite{girshick2015fast} the object proposals are only used to guide the task of spatial pyramid pooling \cite{he2014spatial}. More specifically, under fast R-CNN, much of the convolutional layer computations are pooled and reused. The proposed pooling strategies allow for the extraction of deep features to become a much smaller fraction in the compute time (less than 1s). Using the original convolutional layer output for feature pooling along the proposed windows (object proposals) significantly improves on the complexity of the object detection task. This, for small to medium size images, is shown to significantly reduce the computational complexity of fast R-CNN. On the other hand, in real world applications such as unmanned vehicles, video surveillance and robotics, there is an essential need for object detection in extremely large and high-resolution images. The challenge here is that for high-resolution large images, the initial pruning step for object proposal generation does not scale well. The main contribution of our work is to address the viability of fast R-CNN as an object detection algorithm for high resolution images. 

There is a growing literature on efficient object proposal algorithms, such as BING\cite{cheng2014bing}, EdgeBoxes\cite{zitnick2014edge} and MultiBox\cite{erhan2014scalable}. However, all these algorithms suffer from a significant scaling of the computation with the size of an image (in pixels). More precisely, the processing power required for the pruning phase grows quickly with the size of the image, as the number of rectangular regions in an image is $O(m^2 n^2)$ for an m by n image. As a result, the proposed fast R-CNN framework cannot be directly applied beyond the usual large scale datasets such as Pascal and ImageNet (with image sizes around $500 \times 500$).

In this paper we focus on high resolution images with small objects. We note that, in many practical scenarios of interest, the number of objects of interest does not grow nearly as fast as that of the size of the pixels and the potential bounding boxes. Furthermore, the information in many object proposals are highly correlated, i.e. redundant, due to overlapping. This suggests the possibility of designing efficient object detection schemes which take advantage of the existing sparsity and redundancy. In particular, we design heuristics based on the following attributes of the problem:

\begin{itemize}
\item Overlapping regions share visual features. It is possible to infer reliably about the contents of a region if features from sufficiently overlapping regions are available.
\item Images often exhibit hierarchical structures that are preserved through a reduction in resolution. Observing features from large entities that can be clearly seen at a low resolution could strongly indicate the existence of a smaller high resolution entity that is blurry at the lower resolution.
\end{itemize}

It is easier to illustrate the intuition behind these claims by considering the following example of locating a car in a scene: detecting an image of an engine cover tells any human observer not only the existence of the car in a large region, but also the particular neighboring regions that contain other segments of the same car. On the other hand, in the case of searching for a small car in a large scene, an effective strategy is to first look for large entities such as highways that are likely to contain cars. 

Capitalizing on these intuitive attributes of the problem, we incorporate in any object proposal scheme the following design principles: 1. For any initial region that is likely to be in the neighborhood of an object of interest, make local bounding-box predictions to adjacent objects. 2. Search for regions that are likely to contain one or more small (relative to the region) object(s), then perform detailed processing at the interior of these regions. 

In this paper we introduce a simple approach that combines the two principles using DCNN-based heuristics. Our contributions are:

\begin{itemize}
\item Propose a framework that makes current object detection algorithm more efficient by incorporating the two principles.
\item Train a neural network model called Spatial Correlation Network (SC-Net) on top of DCNN features. The output of this model are heuristics corresponding to the two principles: bounding box predictions and zoom indicators. 
\item Demonstrate the effectiveness of our approach on a high-resolution image subset of the SUN 2012 dataset.  
\end{itemize}

We will present our algorithm in Section \ref{sec:design}. The implementation details used in our experiments are presented in Section \ref{sec:imp}. In Section \ref{sec:exp} we present our empirical results. 

\section{DESIGN OF THE ALGORITHM}
\label{sec:design}
In this section we introduce the design of our algorithm. The section starts with a discussion of the roles of each algorithmic building blocks. The pipeline of our framework is then explained. The section is concluded with a discussion of existing works that are related to our algorithm. 

\subsection{Basic building blocks}
Our algorithm uses several components described below as its basic building blocks. A few of these components are proposed in the literature, while the last two components are specific to our design. In particular, we utilize the following techniques in the literature: deep convolutional neural networks \cite{krizhevsky2012imagenet}\cite{simonyan2014very}\cite{szegedy2014going}, the region feature pooling \cite{he2014spatial}\cite{girshick2015fast}, and object proposers based on low complexity features \cite{cheng2014bing}\cite{zitnick2014edge}\cite{uijlings2013selective}. We also introduce two newly developed components: bounding box predictions and zoom indicators. We will first introduce the existing components.

\begin{itemize}
\item \textbf{Deep convolutional neural network}
The input of the deep convolutional neural network (DCNN) is a color image with arbitrary size. The output is a feature image used to encode high-level visual information for different parts of the input image. The feature image is the output of the last convolutional layers for a DCNN-based image classification model, such as AlexNet \cite{krizhevsky2012imagenet}, VGG \cite{simonyan2014very} and GoogLeNet \cite{szegedy2014going}.

\item \textbf{Region feature pooling}
Region feature pooling converts a sub-region in the feature image into a fixed-length feature vector that describes the underlying sub-region in the input image. One successful technique for this task is the spatial pyramid pooling \cite{he2014spatial} approach. In our algorithm we use a simplified version called RoI pooling which has been deployed as part of the Fast R-CNN \cite{girshick2015fast}.  

\item \textbf{Object proposer (with small fixed size inputs)} In this paper object proposer refers to a conventional object proposal algorithm \cite{uijlings2013selective}\cite{cheng2014bing}\cite{zitnick2014edge}\cite{erhan2014scalable} that proposes potential tight bounding boxes for objects (object proposals) based on the content of the image. While there is a wide variety of object proposers with acceptable performance for processing small to medium size images, their run time to process a large, high resolution images grows quickly unsustainable. To control this complexity, we restrict the input to the object proposer to small images. When the input image (sub-image) is larger than a fixed small size down-sampling is performed. Another class of object proposer we consider is coarse sliding windows generated independent of image contents. We are particularly interested in the performance of this light-weight approach because compared to common object proposer it introduces essentially no overhead to the detection pipeline. 

\end{itemize}

We now discuss the bounding-box predictions and the zoom indicators. They are novel procedures designed to instantiate the two principles we identify in the introduction. These are special purpose neural networks designed to fully utilize the spatial correlation structures of an image.

\begin{itemize}
\item \textbf{Zoom indicator} Algorithmically, the zoom indicator is generated by a procedure that takes as its input a RoI and the DCNN feature image and outputs a scalar in the unit interval. The zoom indicator is used to focus high resolution processing procedures to sub-regions in the image. A region is worth zooming if it is likely to contain small objects. As an efficient strategy to deploy processing power, our algorithm select a small number of sub-regions based on the corresponding zoom indicators. 

\item \textbf{Bounding-box predictions} Bounding-box prediction is useful when we have a region that partially overlaps with an object. Bounding-box prediction uses the features pooled from the initial regions to predict a set of regions that overlaps with the adjacent objects best. Algorithmically it takes as its inputs a RoI (regions-of-interest) and the corresponding DCNN feature image. At its output is a set of adjacent bounding boxes (each adjacent bounding boxes is identified by the coordinates of its top-left and bottom-right corners relative to the input RoI). These outputs are functions of a pooled region feature vector corresponding to the input RoI.

\end{itemize}

We note that although these two components perform conceptually different computational tasks, algorithmically the form of their input is identical. We utilize this fact in our implementation by training a Spatial Correlation Network (SC-Net) to jointly perform theses tasks and output both zoom indicator and bounding box prediction for a given input region. We will discuss the SC-Net and its use case in the proposed pipeline in Section \ref{sec:imp}. 

\subsection{Pipeline of the algorithm}
We first define the input and output of the proposed method before introducing its pipeline. The input of our algorithm is a color image of arbitrary size. We assume the most common inputs are large high resolution images. The output of the algorithm is a set of rectangular boxes $b_t = (x_1^t, y_1^t, x_2^t, y_2^t)$ in the input image (the tuples $(x_1^t, y_1^t), (x_2^t, y_2^t)$ are the coordinates of the top-left, bottom-right corners of the box, respectively), each of which is a proposal for an object of the interested category.

\begin{figure*}[thpb]
      \centering
      \parbox{7in}{\includegraphics[height=2.5in, width=7in]{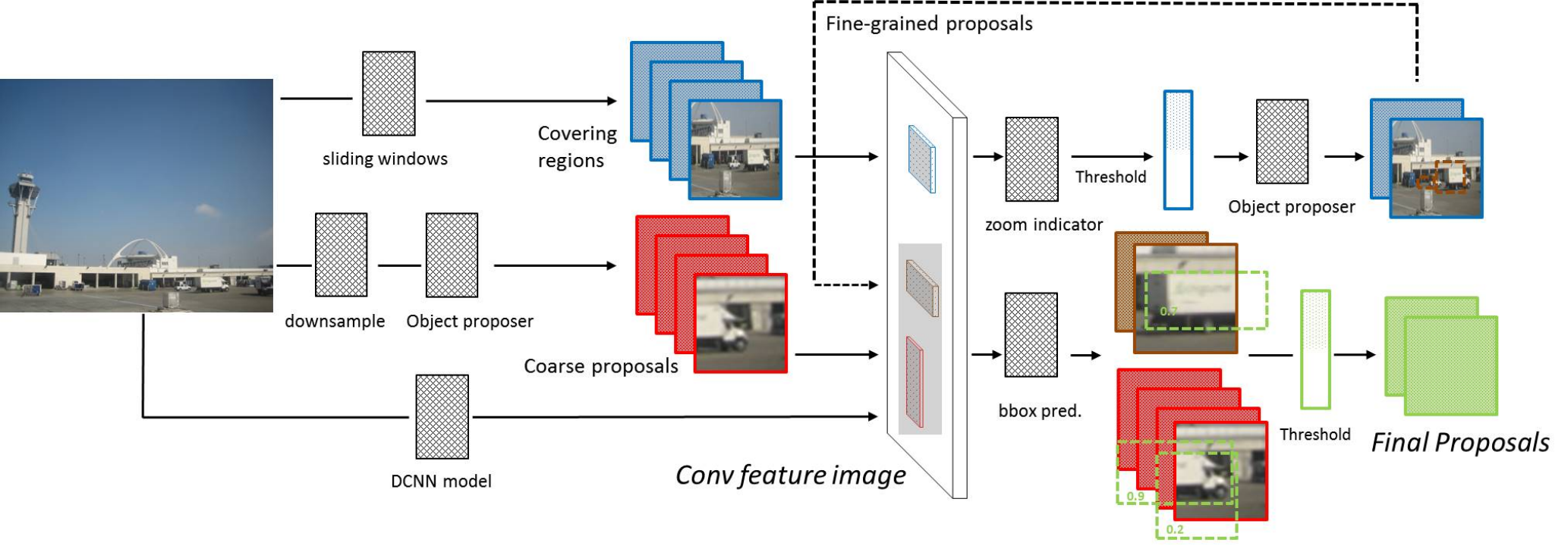}
      }
      \caption{Illustration of the algorithm with pictorial details. Regions with different meanings are encoded in different colors. The red regions are object proposals from a bottom-up process(selective search). The blue regions are sliding windows that are independent of the contents of the image. The green regions are proposals from bounding-box predictions.}
      \label{fig:pipeline_details}
   \end{figure*}

Illustrations of the pipeline of our algorithm is shown in Figure \ref{fig:pipeline_rough} and \ref{fig:pipeline_details}. As the first step, our algorithm computes the DCNN feature image and save it as a global variable for later processing. Along one sequence, the entire image is downsampled to a small fixed size, which will be used as the input to a coarse object proposal algorithm (object proposer). The output of this coarse object proposal sequence is a set of sub-regions, let us denote this set as $A$. Since in this coarse object proposal process a down-sampled version of original image is used, set $A$ is often missing bounding boxes for small objects. A parallel sequence of operations is proposed to address this. Specifically a fairly small cover of this image is extracted from the input image. Each region in the cover plays the role of a potential candidate for a ``zoom-in" operation as follows. For each region in the cover, a region feature vector is pooled and subsequently a zoom indicator is computed. The zoom indicator is designed to identify the regions in the cover that are likely to hold small objects and hence are worth a further high-resolution processing step. Each region with sufficiently large zoom indicator is input to an object proposer that outputs a confined set of sub-regions as additional candidate proposals, we denote this set as $B$. The union of $A$ and $B$ is used as input to the bounding box prediction procedure. This procedure uses the extracted features inside the input regions to output a set of final proposals (denoted as $C$), which is the output of our algorithm. For a complete object detection pipeline, each of the final proposals are then fed into an object category classifier (optionally with the traditional bounding box regression) and subsequently the post-processing procedure for removal of multiple detections. 

\subsection{Related work}
Compared to object proposal algorithms based on low-level bottom-up processing, such as segmentation \cite{uijlings2013selective} and edge detection \cite{zitnick2014edge}, our algorithm utilizes redundancy in the images by modeling the high-level visual concepts explicitly. This strategy seems to be complementary to the low-level approach, which as we will demonstrate does not scale well in high resolution settings. We note that while in our implementation we have chosen specific algorithms, our proposed design can work in companion with a traditional object proposal algorithm to improve its scalability.

Some recent object proposal algorithms are based on a neural net model. For example, the Multi-box algorithm \cite{erhan2014scalable} uses a single evaluation of a deep network to predict a fixed-number of object proposals. This algorithm similarly models high-level visual concepts and benefits from GPU acceleration. However, we note that one crucial detail that prevents an excessive growth in complexity of Multi-box is the use of a carefully designed set of anchor regions. The robustness of this technique in high resolution images containing small objects is unclear. In this light, our algorithm offers to provide a framework that could boost the performance of Multi-box in high resolution setting without significant efforts in domain adaptation. This is an area of future exploration.  

The bounding box prediction method we propose is related to the bounding box regression approach introduced in \cite{girshick2014rich}. The traditional bounding-box regression used in fast R-CNN predicts one bounding box for each class. The assumption is that the spatial support of the RoI overlaps with the underlying object well enough for accurate object category prediction. The regression serves to provide a small correction based on the typical shapes of objects of the given category to get an even better overlapping. In our application a typical input region is assumed to have a small partial overlapping with the object. Our strategy is to focus on the spatial correlation preserved by the geometry of overlapping. We will discuss more about these in the next section. 

\section{IMPLEMENTATION}
\label{sec:imp}
We have implemented our algorithm and tested its effectiveness on image datasets. In this section we discuss the details of the implementation choices in our design. At the core of our method is the pipeline described in Section \ref{sec:design} and the corresponding building blocks described in the same section. In this section we will first discuss the implementation choices for each of the building blocks. After that components that are unique to our approach are discussed in details. 

\subsection{Design Choices}
In our implementation, the deep convolutional neural network model of choice is the AlexNet \cite{krizhevsky2012imagenet} model trained on ImageNet. We note that our algorithm will also work with other more accurate but computationally more expensive pre-trained DCNN models, such as VGG \cite{simonyan2014very} and GoogLeNet \cite{szegedy2014going}\cite{szegedy2014scalable}. 

For the region feature pooling procedure we use an existing algorithm called RoI pooling, as described in \cite{girshick2015fast}. This algorithm is essentially a simplified version of the spatial pyramid pooling algorithm \cite{he2014spatial}. The RoI pooling is adopted for the availability of efficient implementation. 

We mainly test two object proposers: the selective search \cite{uijlings2013selective} algorithm and coarse sliding window search. The former is a object proposer based on low complexity features and bottom-up segmentation. The latter is a content independent mechanism. We will provide more details in the later part of this section. 

In our implementation the bounding box predictions and zoom indicators are obtained by a jointly designed and trained Spatial Correlation Network (SC-Net). We will discuss the implementation details of the SC-Net below. 

\subsection{Object proposers}
One of the object proposer we investigate is the selective search algorithm, since it is the pruning algorithm adopted in the benchmark fast R-CNN algorithm \cite{girshick2015fast}. To control the complexity of selective search, as a parameter of the algorithm we set the maximum input size and downsample the image if it exceeds that size. Setting a small size negatively affects the quality of the proposals due to loss in resolution (see Figure \ref{fig:runtime}). In our experiments, we change this parameter to investigate the runtime required for the algorithm to achieve various level of accuracies. 

Another object proposer we investigate is to blindly generate a coarse set of sliding windows. We note that this can also be viewed as a simple object proposer. It is a sensible approach in this context since bounding box prediction can adjust boxes with partial overlapping with objects. A crucial detail is that the sizes of the windows have fixed ratios to the size of the region under consideration. Thus when deployed to a small sub-region small objects inside that region can be recovered.

\subsection{Implementation of SC-Net}
While the zoom indicator and bounding box predictions are performing conceptually different tasks, they build on the same RoI feature input. As mentioned earlier in our design we utilize this to implement both sets of outputs in a single network (SC-Net). The advantage of this approach is that it reduces the number of parameters to be trained and improves computational efficiency at inference time. The adoption of a single neural net also simplifies the training procedure.

The SC-net takes as input a region. It first pools a fixed-length feature vector from the corresponding sub-region in the convolutional feature image. The outputs, a function of this feature vector, are $K$ bounding-box coordinates and their associated confidence scores in addition to the zoom indicator $u \in \mathbb{R}$ that describes the likelihood that a small object is present at the interior of the input RoI. 

The bounding-box prediction decides whether an input region overlaps with an object non-trivially and at the same time predicts a bounding box for that object. In our design however, we output $K$ such predictions for each input region. The $K$ predictions are each trained to respond to one particular overlapping pattern between the input RoI and the object. We heuristically define $K=13$ such categories. The detail of the definition can be found in the Appendix for interested readers. We note that this approach in effect provide a separate set of parameters for each of the heuristically defined overlapping pattern. It helps in providing stronger training signal (especially for the coordinate predictions) so that the training data is used more efficiently. 

\subsubsection{Network architecture and I/O}
\begin{figure}[thpb]
      \centering
      \parbox{3in}{\includegraphics[height=1.5in, width=3.2in]{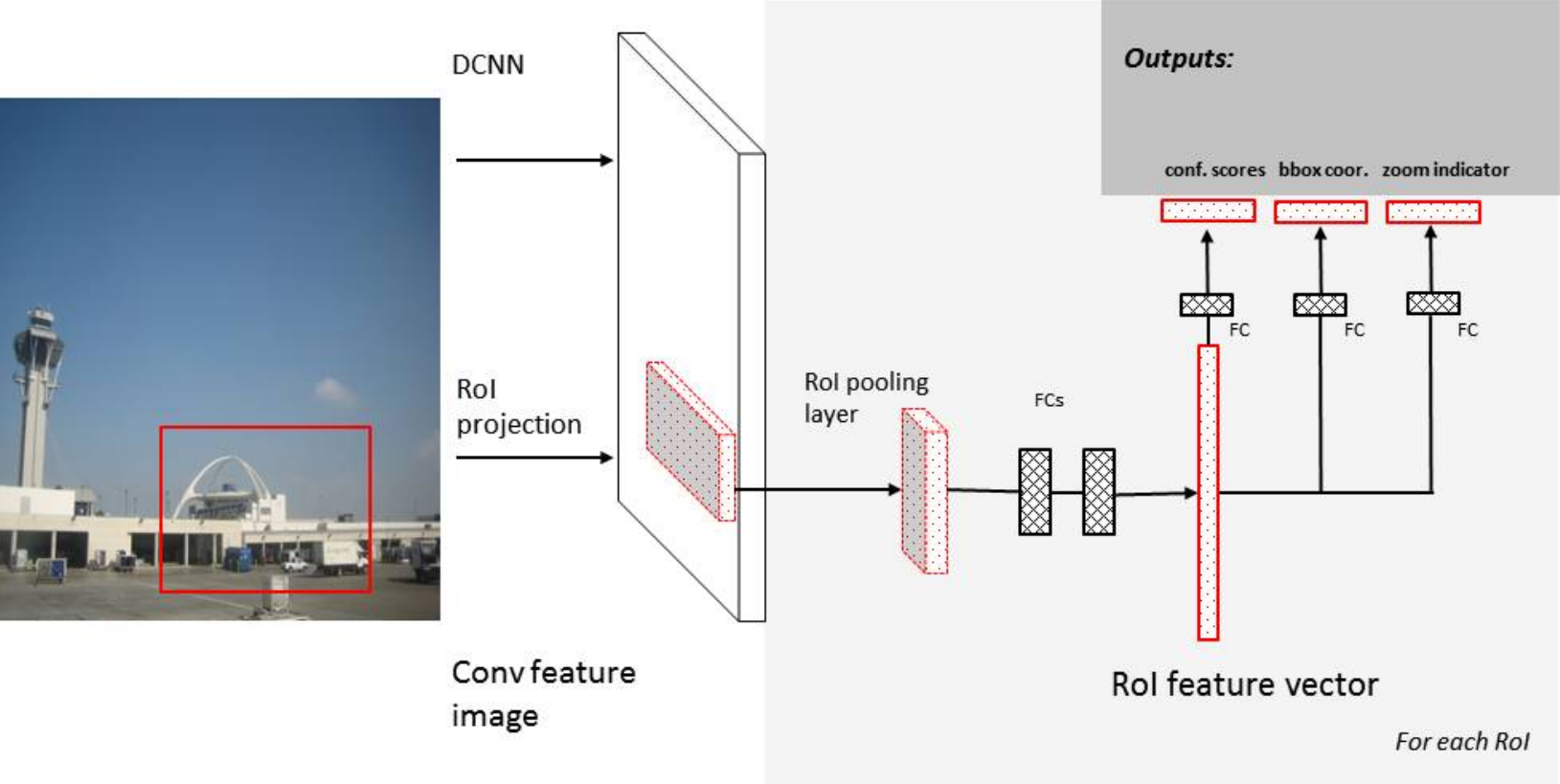}
      }
      \caption{The architecture of our the Spatial Correlation Network (SC-Net).}
      \label{fig:active_net}
 \end{figure}

The shaded part in Figure \ref{fig:active_net} illustrates the SC-Net. The pooled features vector is $9216$ dimensional. The vector is fed into two 4096 dimensional fully connected (fc) layers. The output of the last layer is the 4096 dimensional RoI feature vector. A single fully connected layer is applied for each of the three output components to obtain outputs for the bounding-box prediction networks and zoom indicator networks. The activation function for both the zoom indicator and the confidence scores are the sigmoid function. The one for the bounding box deltas is the identity function. 

\subsubsection{Training data}
We use images in the training set of SUN 2012 detection dataset that contain either ``car" or ``person". As part of the ground truth annotations, tight bounding boxes for objects in these two categories are provided alongside with their class labels. Since the original dataset contains very fine-grained labels, to provide more training (and correspondingly testing) data we merge visually similar sub-categories, such as ``car occluded", ``person walking" etc., into the two main categories. We augment the training set with horizontal mirror images.

From each training images we extract RoIs that are either object proposals from selective search or ground truth boxes. These RoIs are used as training examples. The training labels are constructed for each training RoI using the available bounding box annotations for the underlying image. The labels for zoom indicator is apparent: we assign label 1 if and only if there is one ground truth object contained inside the input region that is less than 10\% of the region area. For bounding box prediction, each input RoI is paired with its best overlapping ground-truth object and the unique overlapping pattern is determined. At the same time, the intersection-over-union (overlapping) score between the ground-truth and the input RoI is computed. If the overlapping score is above a threshold, the confidence score corresponding to the determined overlapping pattern is assigned label 1, and the corresponding bounding-box coordinate is filled in accordingly. The other coordinates are assigned dummy labels and zero learning weights. 

\subsubsection{Loss function}
We adopt a multi-task loss function similar to the one used in \cite{girshick2015fast}. The loss for the bounding-box coordinates are the smooth-L1 loss. For the confidence score for bounding box predictions and the zoom indicator we adopt the sigmoid cross-entropy loss function commonly used in binary classification. The loss function is minimized using mini-batch stochastic gradient descent. For interested readers more details of the training procedure are provided in the Appendix. 

\subsection{Use of the SC-Net}
\subsubsection{Regions for high resolution processing}
To generate the cover of the image by small candidate regions we utilize a standard sliding windows of a fixed size and stride distance. As shown in Figure \ref{fig:pipeline_details}, the regions are also used as input to the SC-Net. While the SC-Net outputs both zoom indicators and bounding box predictions for each sub-regions, we only use the zoom indicators. In particular, a threshold is set at the zoom indicator to select a smaller set of regions for high resolution processing. 

\subsubsection{Bounding box predictions}
The coarse proposals and the fine-grained proposals (see Figure \ref{fig:pipeline_rough} and \ref{fig:pipeline_details}) are fed into the SC-Net as input RoIs. Again, while both the zoom indicators and the bounding box predictions are available after evaluating the top layers of the network, only the bounding box predictions are used. A threshold is set at the confidence score of the predictions to ensure that only a small set of predictions are kept.

\section{EXPERIMENTS AND ANALYSIS}
\label{sec:exp}
In this section we show empirical results of our approaches against the two baseline approaches: sliding window and fast R-CNN. We will first introduce our evaluation methodology, in particular the dataset and the metric we adopted. Then we will present the comparison of our approaches against the baseline. To understand the relative contributions of the two strategies, we show the performance of our algorithm as the design components are incrementally turned on. This section is concluded with a discussion of the advantages and limitations of our method (supplemented with visual illustrations) that points to future directions. 

\begin{figure*}[thpb]
      \centering
      \parbox{3in}{\includegraphics[height=1.6in, width=3in]{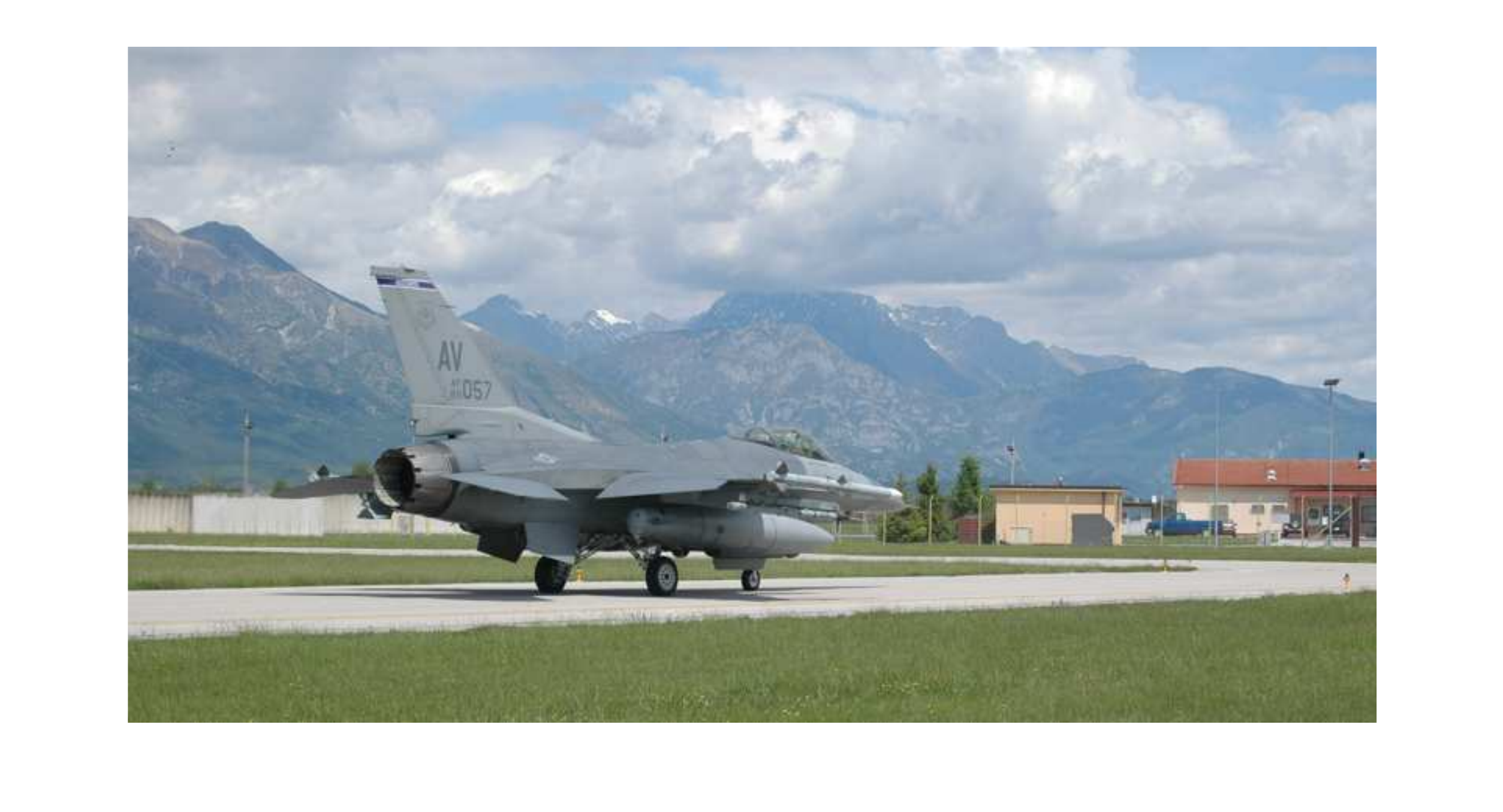}
      }
      \parbox{3in}{\includegraphics[height=0.8in, width=1.5in]{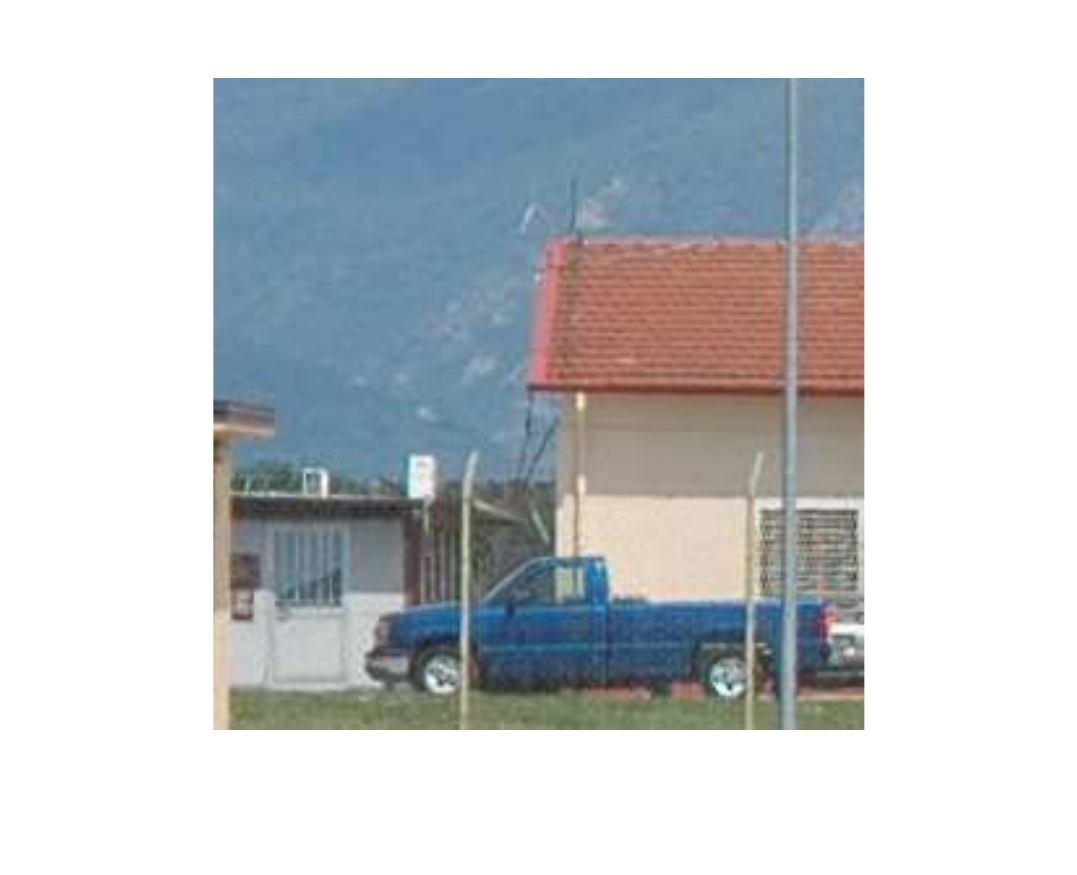}\includegraphics[height=0.8in,width=1.5in]{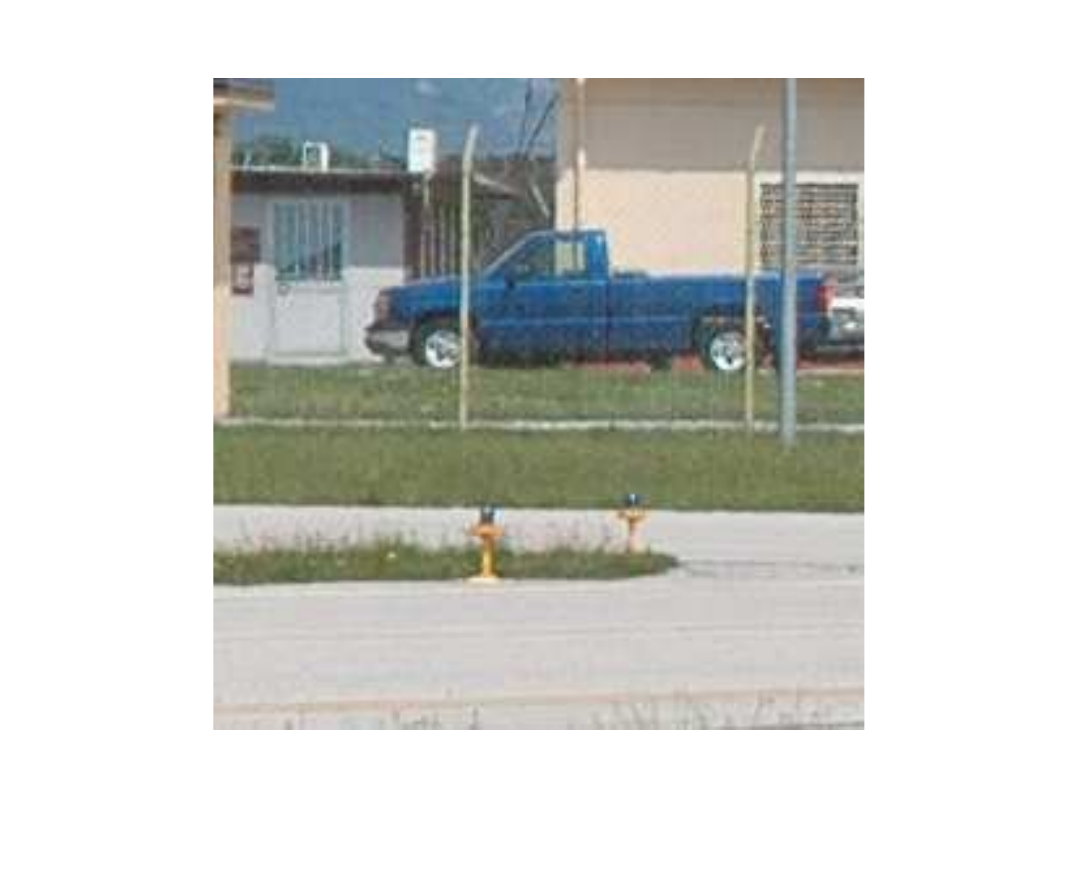}
      \includegraphics[height=0.8in, width=1.5in]{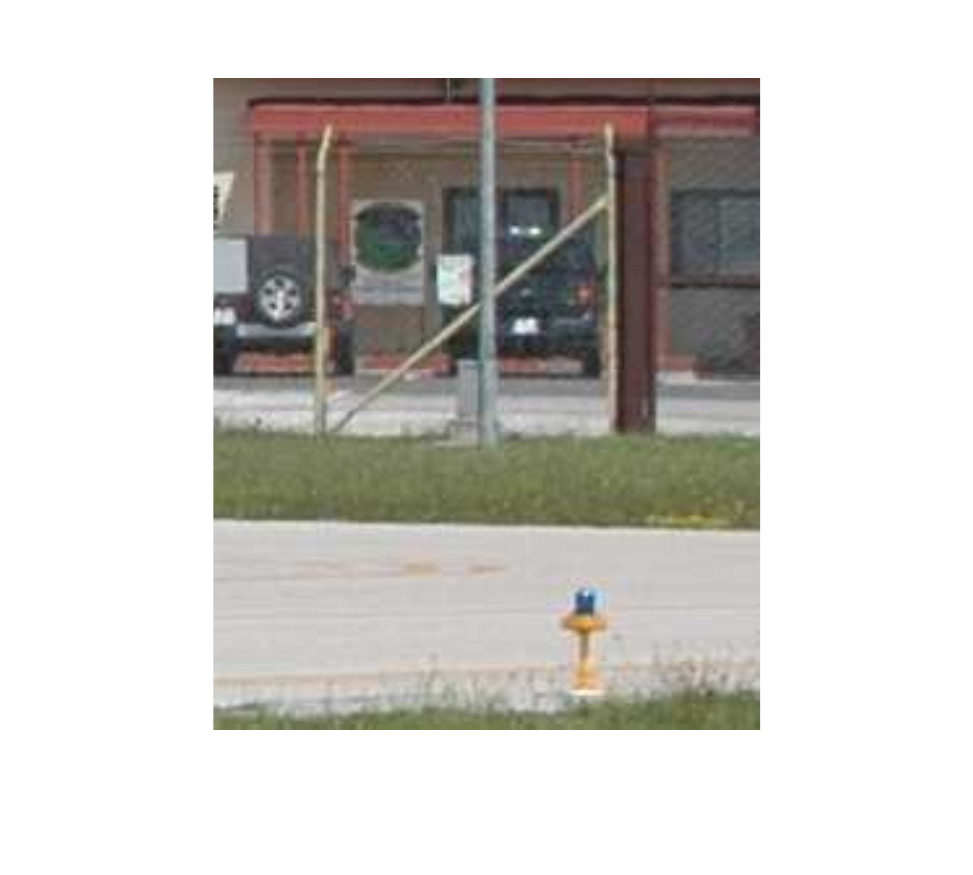}\includegraphics[height=0.8in, width=1.5in]{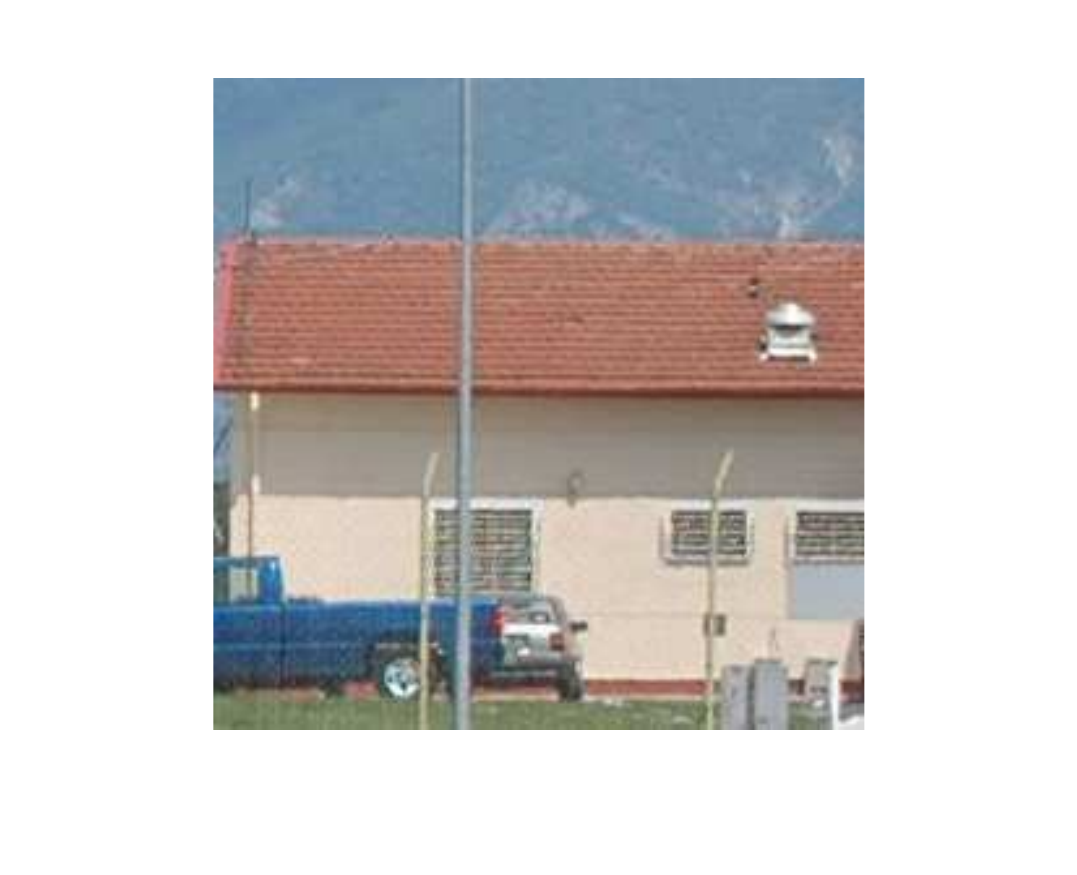}
}
      \parbox{3in}{\includegraphics[height=1.6in, width=3in]{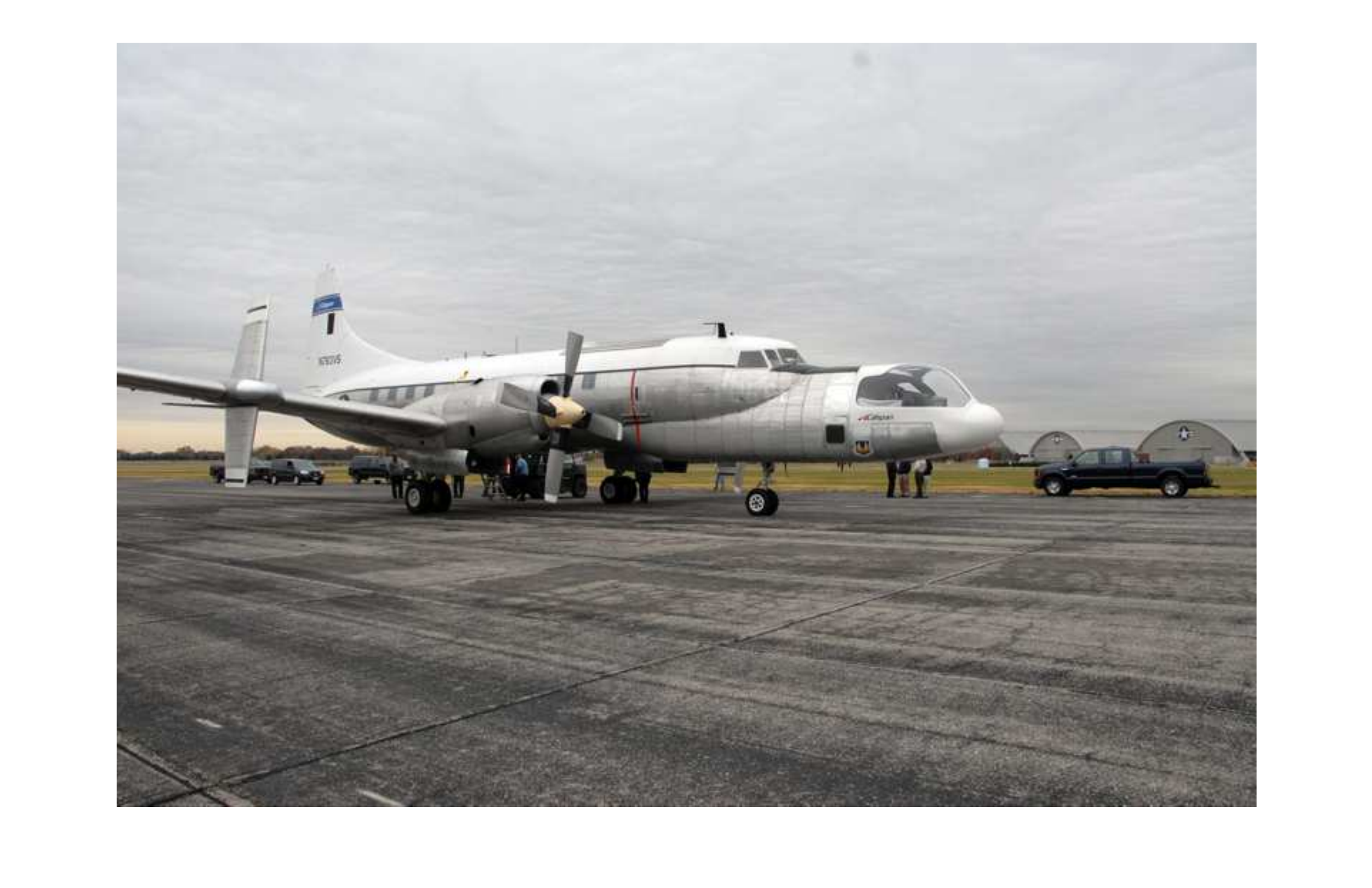}
      }
      \parbox{3in}{\includegraphics[height=0.8in, width=1.5in]{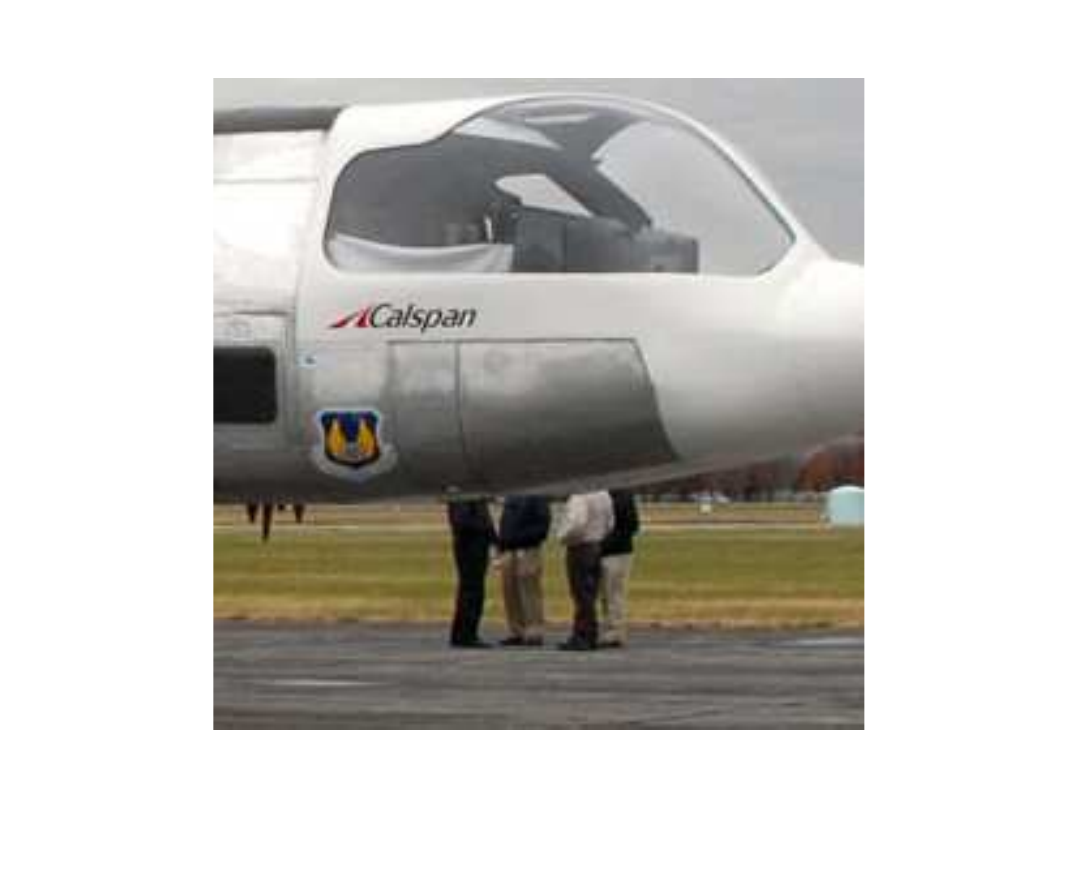}\includegraphics[height=0.8in,width=1.5in]{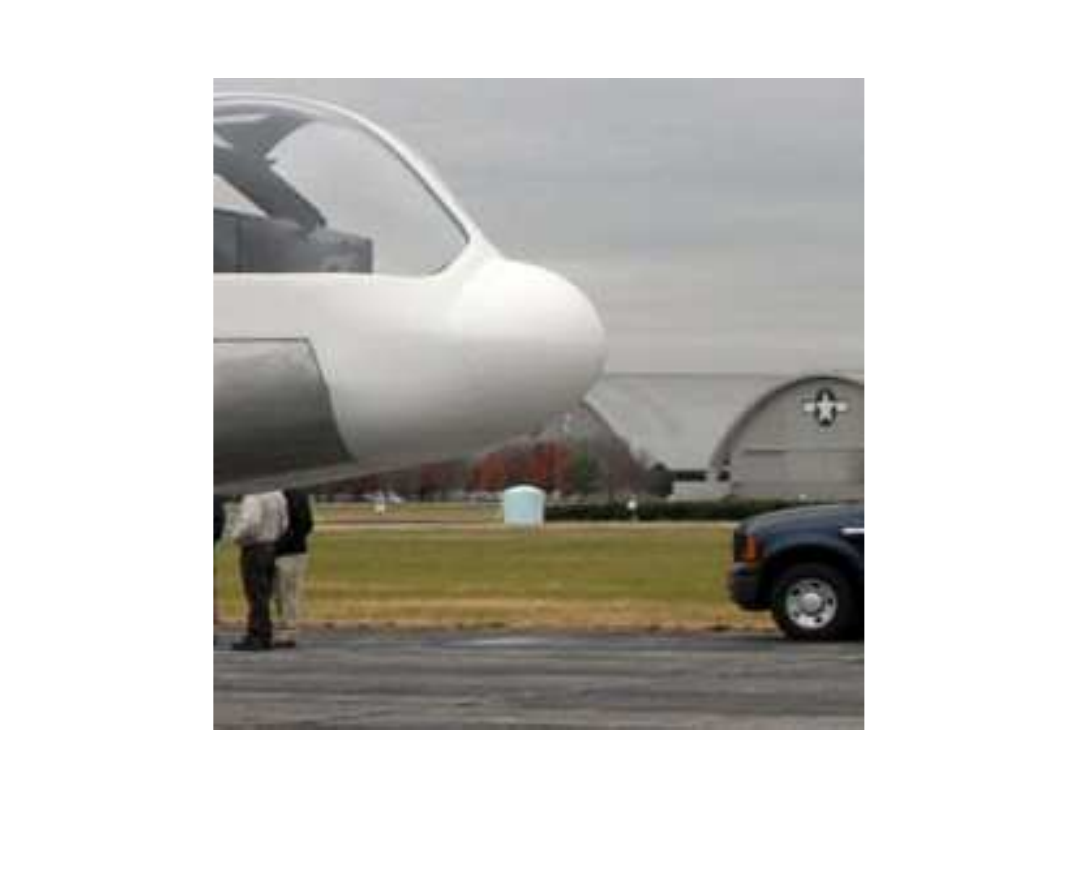}
      \includegraphics[height=0.8in, width=1.5in]{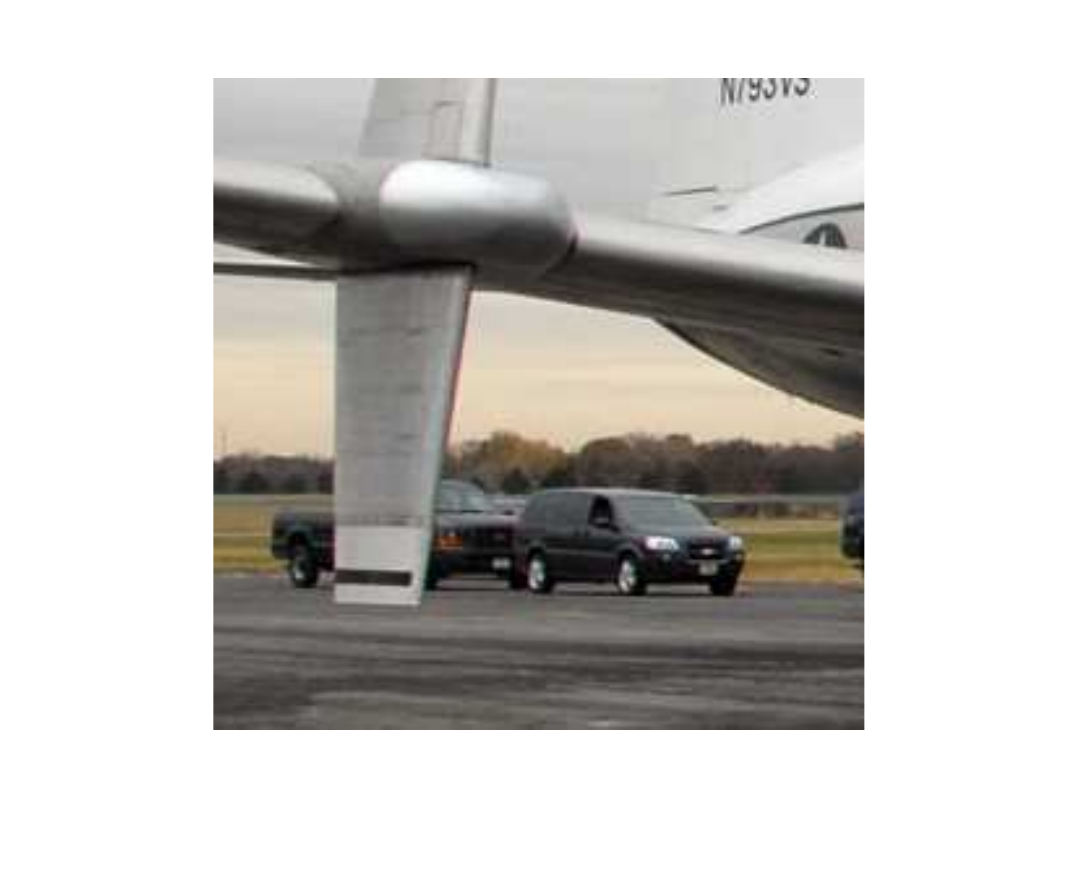}\includegraphics[height=0.8in, width=1.5in]{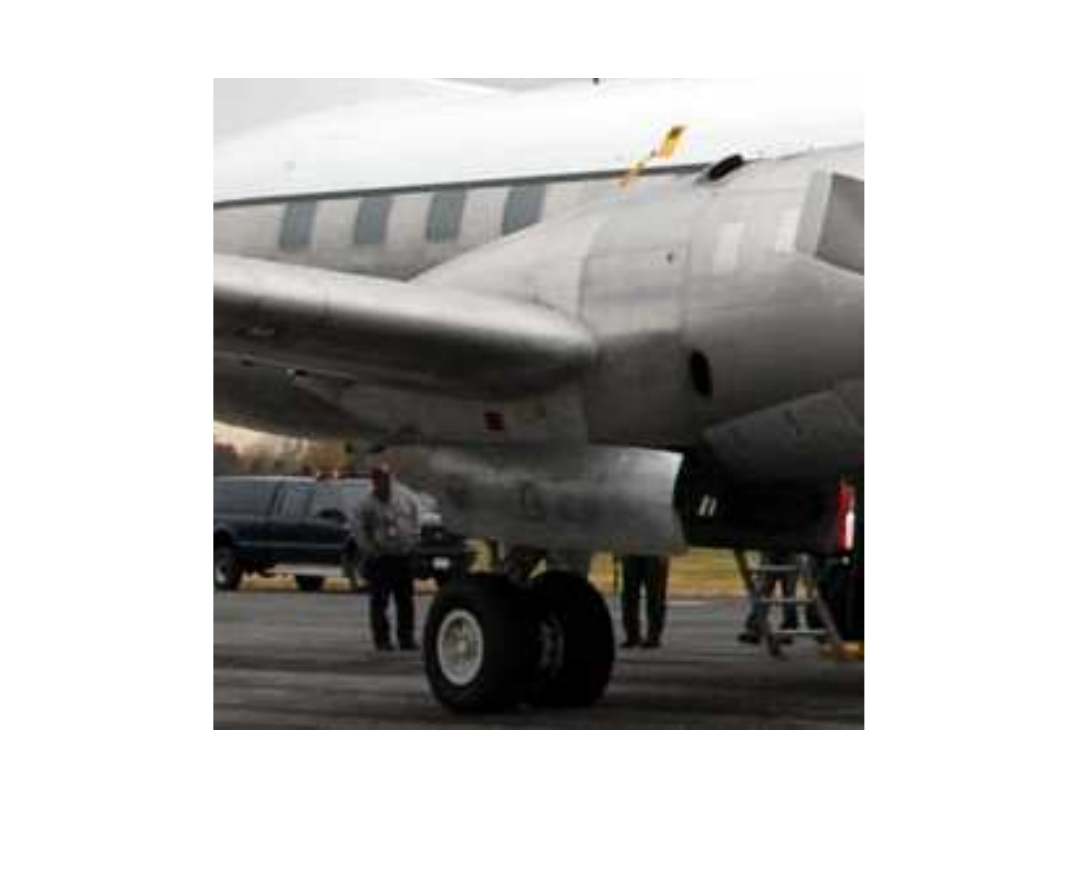}
}
      \caption{The regions selected for high-resolution processing. The left column shows the original image. The right column shows the four regions with highest zoom indicator values. It is clearly seen that regions containing small cars or persons are selected from the process.}
      \label{fig:search}
   \end{figure*}

\subsection{Evaluation methodology}
The evaluation is performed on a subset of the test set of SUN 2012 detection dataset \cite{xiao2010sun}. The subset consists of 176 images, each with a shortest side of at least 1200 pixels. All included images contain object instances in one or both of the two categories: car and person. We merge fine-grained labels that are visually similar to the two main categories using the same procedure we adopted to process the training set. 
 
We choose to evaluate the performance of the approach by plotting out its runtime against reliability. The reliability metric of choice is the recall metric, which is used widely to evaluate the accuracy of object proposals \cite{hosang2014good}. It is defined as the percentage of ground truth boxes that can be assigned a bounding box with sufficiently large overlapping (intersection over union score greater or equal to $0.5$). An algorithm is more efficient if it achieves the same recall at a smaller runtime.

\begin{figure}[thpb]
      \centering
      \parbox{3in}{\includegraphics[height=2.3in, width=3in]{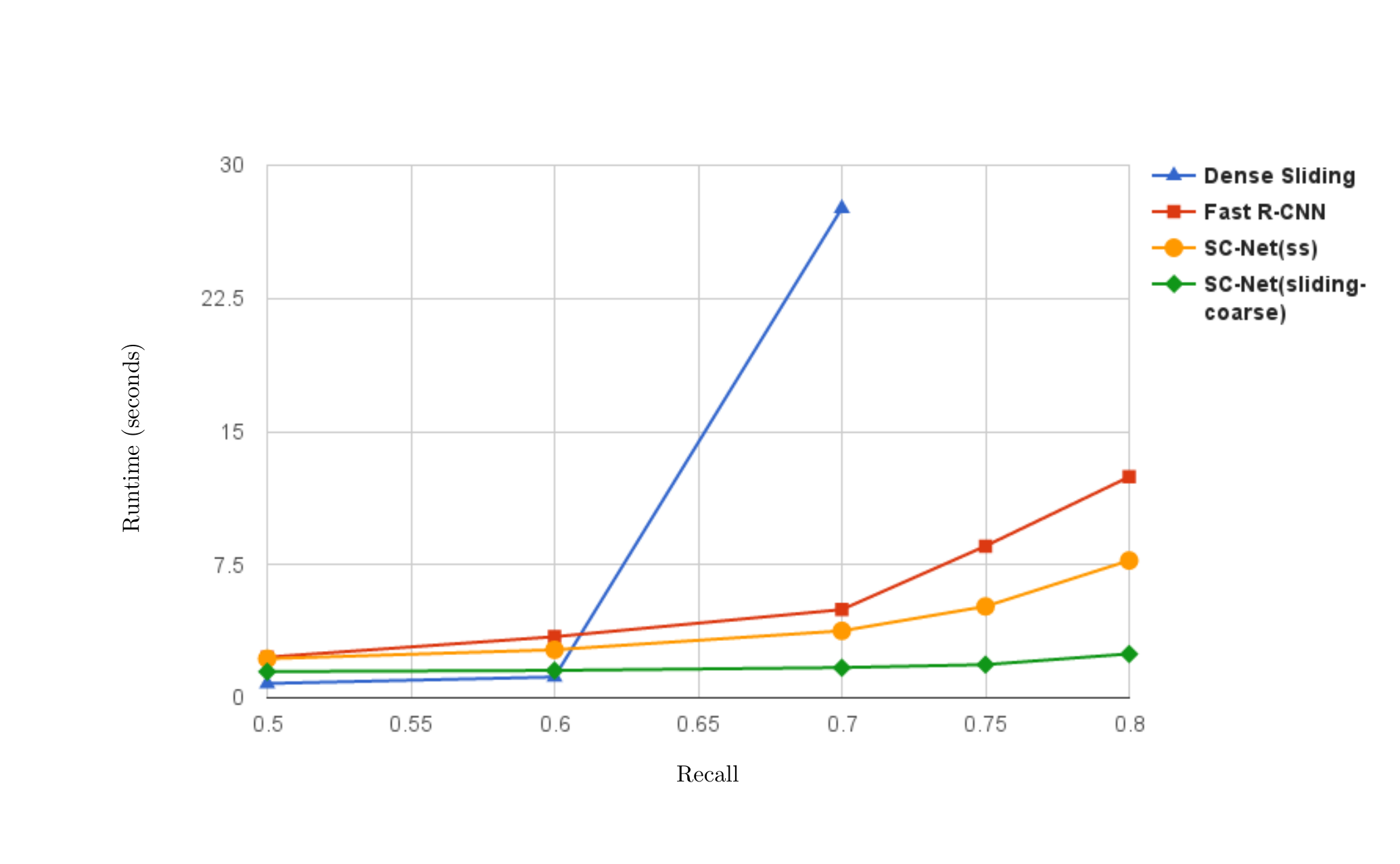}
      }
      \caption{Comparison of runtime (in seconds per image) required to achieve different levels of average precision.}
      \label{fig:runtime}
   \end{figure}

\subsection{Performance comparison}
For all our experiments we feed the object proposals to region classifiers trained in the same procedure as in Fast R-CNN but on our customized training dataset. We compare the accuracy of the bounding boxes after adjustments made by bounding box regression. We note that this provides a fair comparison as all the methods we compare utilizes the accurate DCNN features. The runtime is for the complete pipeline including the classification steps. 

\subsubsection{Benchmark comparison}
We compare the performance of the following settings. 

\begin{itemize}
\item \textbf{Dense sliding window} Apply classifier to a dense set of sliding windows. The boxes are adjusted by bounding-box regression by the region classifier.
\item \textbf{Fast R-CNN} Replace the dense sliding windows with proposals generated by objective search with different input resolution. This is essentially the Fast R-CNN pipeline.
\item \textbf{SC-Net (ss)} The algorithm as illustrated in Figure \ref{fig:pipeline_rough} and \ref{fig:pipeline_details}. Both the coarse proposals and the fine-grained proposals are generated using selective search with property sampled input images. 
\item \textbf{SC-Net (coarse sliding)} Replace the object proposer used in SC-Net (ss) with a coarse set of sliding windows. The sizes of the windows have fixed ratio to the size of the input region. 
\end{itemize}

Figure \ref{fig:runtime} shows the comparison. As expected, the dense sliding window approach is very inefficient, even with the bounding-box regression. The Fast R-CNN suffers from the slow runtime of the selective search algorithm when recall is high. Our approaches based on the SC-Net model clearly shows advantages in runtime, especially at high recall points. We note that the SC-Net (coarse sliding) approach offers best trade-off between complexity and reliability.

\begin{figure}[thpb]
      \centering
      \parbox{3in}{\includegraphics[height=2.3in, width=3in]{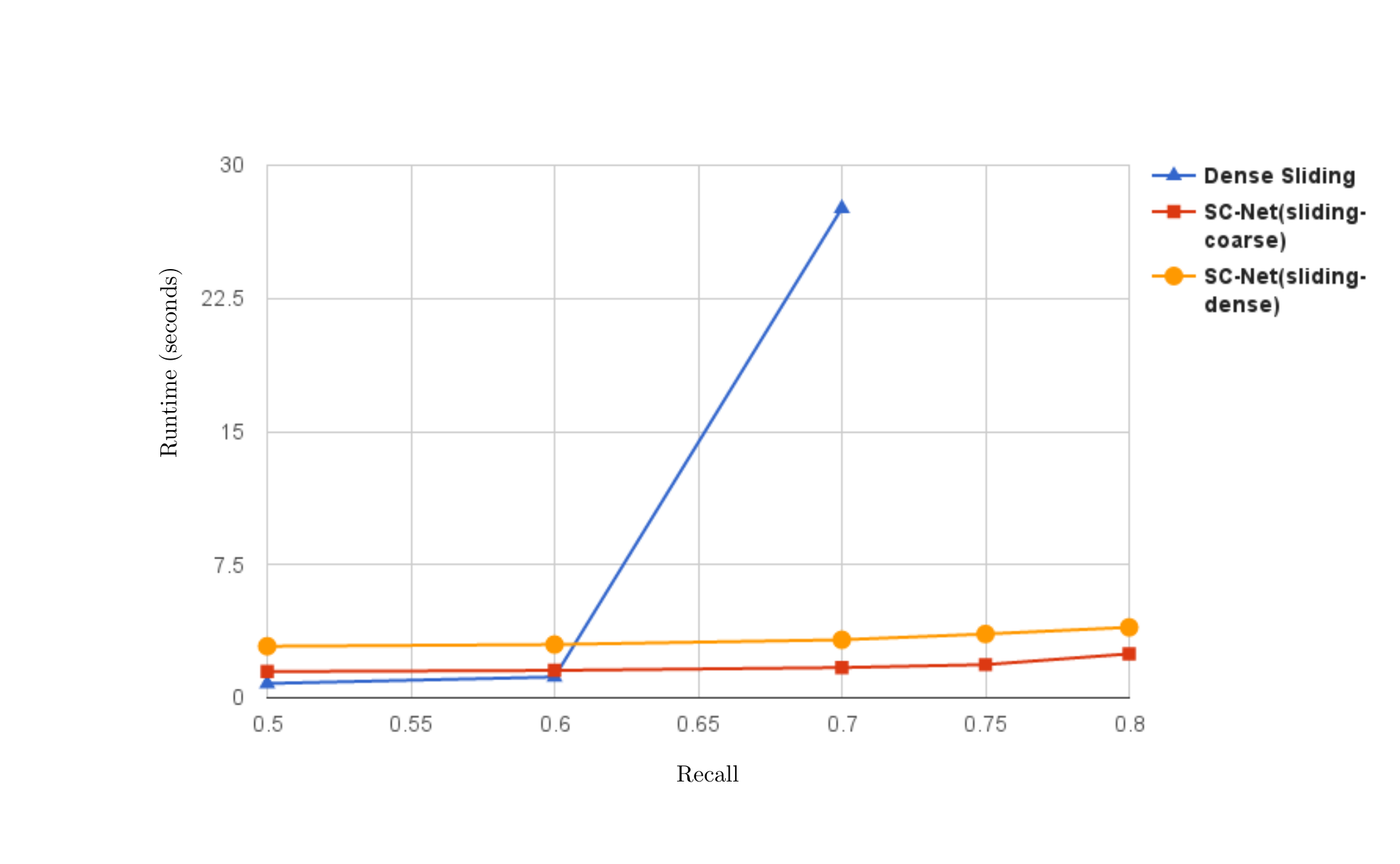}
      }
      \caption{Contribution of the two strategies.}
      \label{fig:breakdown}
   \end{figure}

\subsubsection{Contribution of design components}
To further understand the contribution of the design components, we compare three cases: dense sliding window, SC-Net (coarse sliding) and SC-Net (dense sliding). The SC-Net (dense sliding) approach uses the bounding-box prediction. It is different from SC-Net (coarse sliding) in that it uses a dense set of windows rather than the two-step process of applying coarse windows. In this way, the difference of dense sliding window and SC-Net (dense sliding) is the gain due to bounding box prediction. The difference between SC-Net (dense sliding) and SC-Net (coarse sliding) is the gain due to zoom in processing. The comparison is shown in Figure \ref{fig:breakdown}. It is evident that both strategies are essential for efficient object detection.

The effect of zoom in processing can also be seen from the visualization provided in Figure \ref{fig:search}. Since regions that contain small objects are assigned high zoom indicators, they are processed in finer details which allows the small objects to be recovered. This strategy is more efficient than SC-Net (dense sliding) since small boxes inside unpromising sub-regions are never processed.

\section{CONCLUSIONS}
In this paper, we propose an effective algorithm to perform object detection for high resolution images. Our approach utilizes two complementary spatial correlation structures that exists in images: correlation that stems from adjacency of regions, and correlation that stems from hierarchical structures of features . This is achieved by jointly train a Spatial Correlation Network on top of deep convolutional neural network models pre-trained on large image datasets. Through experiments on SUN 2012 dataset we demonstrate the efficiency of our algorithm in processing high resolution images. 

We note that there are some important future directions both in theory and practice. From a practical perspective, an efficient implementation of the object proposer that can fully utilize the sparsity structure revealed by high level image features (zoom indicators) could improve the computational efficiency further. The gain of utilizing more than one step of zoom in is yet to investigate. We also demonstrate how bounding box prediction can make sliding window proposer very effective, which suggests a good strategy for time sensitive applications. From a theoretical perspective, the heuristics adopted in this work, effective as they are, are biased towards the authors' observations on the visual world and might be sub-optimal. A systematical extension that allows the system to identify and utilize the rich redundancy structure of the visual world is an important future direction. 

\section*{ACKNOWLEDGMENT}
We would like to thank our collaborators Daphney-Stavroula Zois, Maxim Raginsky and Svetlana Lazebnik for useful discussions and suggestions. 

\bibliographystyle{IEEEtran}
\bibliography{IEEEabrv,root}

\section*{APPENDIX}
\subsection{Definition of overlapping patterns}
The categories are defined along to the following two orthogonal cues. The first cue is region inclusion: the RoI contains the object, the RoI is contained by the object, the RoI overlaps (neither of the former two) with the object (3 categories). The second cue is the relative center location: upper left, upper right, bottom left, and bottom right (4 categories). This quantization is supplemented by a special category that represents an ideally large overlapping between the RoI and the object (greater than 0.7 in overlapping score), making $K=3\times4+1 = 13$.

\subsection{Training details of neural network models}
Training is performed through stochastic gradient descent. The gradient is computed in mini-batch of size 128. The samples in each batch are drawn randomly from two images (each with 64 samples). For coherent training, the overlapping between a RoI and its closest ground truth is considered too small when the overlapping score is less than 0.1. Correspondingly an object with overlapping less than 0.1 is considered small for the zoom indicator. The overlapping is considered good enough when the overlaping score is greater than 0.7. In this case the overlapping pattern belong to the special category that represents an ideally large overlapping between the RoI and the object.

\subsection{Parameters in experiments}
The threshold for bounding-box predictions is 0.001 for SC-Net (ss), and the one at zoom indicators is 0.5. We change the bounding-box prediction threshold for SC-Net (sliding-dense) and SC-Net (sliding-coarse) to get results at different reliability. For coarse sliding windows, the windows are squares with length that are 1/2 and 1/4 of the shorter side of the input image. This is supplemented by squares with length 1/8 and 1/16 of the shorter side of the input image for dense sliding windows. The step size of these windows are 1/4 of their side length. The covering regions are windows with length that are 1/4 of the shorter side of the input image with a step size that is 1/2 of their sides. These windows are subset of the coarse sliding windows applied to the entire image. 

\end{document}